\documentclass[prb,10pt,onecolumn,notitlepage,aps,floatfix]{revtex4-1}
\usepackage[english]{babel}
\usepackage[utf8]{inputenc}
\usepackage{amsmath,amssymb,bm,color,comment,enumitem,graphicx}

\usepackage{float}
\makeatletter
\let\newfloat\newfloat@ltx
\makeatother

\usepackage{algorithm}
\usepackage{algpseudocode}
\usepackage[colorlinks=true,linkcolor=blue,citecolor=blue,urlcolor=blue]{hyperref}

\newcommand{\mX}[0]{{\mathcal{X}}}
\newcommand{\dropcap}[1]{#1}
\newcommand{\mY}[0]{{\mathcal{Y}}}

\newcommand{\mE}[0]{{\mathcal{E}}}
\newcommand{\mM}[0]{{\mathcal{M}}}
\newcommand{\mG}[0]{{\mathcal{G}}}
\newcommand{\mL}[0]{{\mathcal{L}}}
\newcommand{\mP}[0]{{\mathcal{P}}}
\newcommand{\mD}[0]{{\mathcal{D}}}
\newcommand{\mI}[0]{{\boldsymbol{\mathcal{I}}}}
\newcommand{\X}[0]{{\mathbf{X}}}
\newcommand{\F}[0]{{\mathbf{F}}}
\newcommand{\Y}[0]{{\mathbf{Y}}}
\newcommand{\T}[0]{{\bm\Theta}}
\newcommand{\change}[1]{{{#1}}}
\newcommand{\rd}[0]{{\rm d}}
\usepackage{xcolor,listings}
\lstdefinestyle{mystyle}{
    commentstyle=\color{olive},
    keywordstyle=\color{magenta},
    stringstyle=\color{blue},
    numberstyle=\tiny\color{gray},
    basicstyle=\ttfamily\small,
    breakatwhitespace=false,         
    breaklines=false,                 
    captionpos=b,                    
    keepspaces=true,                 
    numbers=left,                    
    numbersep=1pt,                  
    showspaces=false,                
    showstringspaces=false,
    showtabs=false,                  
    tabsize=1
}
\begin{document}
\title{Parameter uncertainties for imperfect surrogate models in the low-noise regime}

\author{Thomas D. Swinburne}
\email[Author to whom correspondence should be addressed: ]{thomas.swinburne@cnrs.fr}
\affiliation{Aix-Marseille Universit\'{e}, CNRS,
CINaM UMR 7325, Campus de Luminy, 13288 Marseille, France}
\author{Danny Perez}
\email{danny\_perez@lanl.gov}
\affiliation{Theoretical Division T-1, Los Alamos National Laboratory, Los Alamos, USA}

\begin{abstract}
Bayesian regression determines model parameters by minimizing the expected loss, an upper bound to the true generalization error. However, this loss ignores model form error, or misspecification, meaning parameter uncertainties are significantly underestimated and vanish in the large data limit. As misspecification is the main source of uncertainty for surrogate models of low-noise calculations, such as those arising in atomistic simulation, predictive uncertainties are systematically underestimated. 
We analyze the true generalization error of misspecified, near-deterministic surrogate models, a regime of broad relevance in science and engineering. We show that posterior parameter distributions must cover every training point to avoid a divergence in the generalization error and design a compatible \textit{ansatz} which incurs minimal overhead for linear models. The approach is demonstrated on model problems before application to thousand-dimensional datasets in atomistic machine learning. Our efficient misspecification-aware scheme gives accurate prediction and bounding of test errors in terms of parameter uncertainties, allowing this important source of uncertainty to be incorporated in multi-scale computational workflows.
\end{abstract}

\date{\today}
\maketitle

\dropcap{S}urrogate models are widely used across science and engineering to efficiently approximate the output of computationally intensive simulation engines
~\cite{alizadeh2020managing,deringer2019machine,lapointe2020machine,nyshadham2019machine,montes2022training,bonatti2022cp,kudela2022recent}.
Domain expertise is leveraged to design optimal features and 
model architecture before parameters are inferred from training data generated by the engines. In a broad range of applications, 
the inference problem shares three key characteristics:
\begin{enumerate}[label=\alph*)]
    \item Simulation engines are near \textit{deterministic}: outputs have vanishing aleatoric uncertainty, e.g. the atomic energy $V({\bf X})$ from some quantum chemistry calculation is a near-deterministic function of atomic positions ${\bf X}$, \change{with very weak uncertainty due to e.g. convergence of the electronic minimization scheme}.
    \item Surrogate models are \textit{misspecified}: no single choice of the $P$ parameters can match all $N$ observations, meaning parameters are intrinsically uncertain.
    \item Surrogate models are \textit{underparametrized}: large quantities of training data are available, such that standard epistemic uncertainty estimates from Bayesian inference vanish, typically corresponding to the regime $P/N\ll1$.
\end{enumerate}

Misspecification affects both finite capacity models and deep learning 
approaches with finite training resources
~\cite{lahlou2021deup,psaros2023uncertainty}.
A misspecification-aware learning scheme should minimize the cross entropy 
between predicted and observed data distributions, known as the 
\textit{generalization error}, or under certain technical definitions, the population risk (see \cite{alquier2021user,morningstar2022pacm}). 
However, the generalization error is typically not tractable for learning,
both due to numerical instabilities and the lack of theoretical bounds for convergence.
Bayesian learning schemes instead minimize an upper bound, the \textit{expected loss} (log likelihood), which ignores misspecification but admits a robust learning scheme. 
Epistemic uncertainties are thus severely underestimated, vanishing in the underparametrized limit, with broad implications for surrogate model selection, uncertainty quantification and error propagation, as noted by multiple groups \cite{masegosa2020learning,kato2022view,lotfi2022bayesian,lahlou2021deup,psaros2023uncertainty,imbalzano2021uncertainty}.
\\

In this paper, we analyze the generalization error of near-deterministic, 
misspecified and under-parametrized surrogate models, i.e. those typically used 
to approximate simulation engines in science and engineering. \change{Our central 
application is to the problem of fitting interatomic potentials for atomic 
simulations\cite{goryaeva2021,montes2022training,li2020complex,wood2018extending}, where the training data is generated by quantum 
chemistry calculations. However, our analysis applies to the fitting of any surrogate model to a near-deterministic ground truth.} Posterior distributions strictly become unbounded as the aleatoric uncertainty vanishes, meaning the low-noise/near-deterministic limit is only taken as a theoretical device to derive conditions that any posterior distribution must obey to avoid a logarithmic divergence in the generalization error of that distribution. We then design an \textit{ansatz} that respect 
this condition and admits a robust learning scheme at finite $N$, with standard 
treatment of epistemic ($N$-dependent) errors, reducing to a constrained loss 
minimization problem for large $N/P$.
Our final misspecification-aware parameter posterior has finite uncertainty even as $N/P\to\infty$, whilst the generalization error of the posterior from Bayesian inference diverges. \change{The approach thus captures misspecification uncertainties missed by standard Bayesian regression schemes.}\\

\change{Our implementation for linear models is a robust and to our knowledge unique means to estimate misspecification uncertainties, and thus is the optimal available choice when fitting high-dimensional linear models to low-noise data.}\\

Our main contributions are:
\begin{enumerate}[label=\alph*)]
    \item We define \textit{pointwise optimal parameter sets} (POPS) for 
    each training point, within which model predictions are exact. Parameter distributions must have mass in every POPS to avoid a divergent generalization error. 
    \item We use the ensemble of loss minimizers from each POPS to design 
    two \textit{ansatz} posteriors, a reweighted $\mathcal{O}(N)$ ensemble 
    and a bounding $\mathcal{O}(P)$ hypercube, both of which respect POPS occupancy 
    and give a finite generalization error, lower than the posterior from Bayesian inference.
    \item For linear models, our \textit{ansatz} can be efficiently evaluated via rank-one updates to a leverage-weighted loss minimizer. This gives an efficient scheme to incorporate misspecification into linear regression, which we show gives minimal computational overhead (approximately a factor of two in training) over Bayesian inference.
    \item \change{A \texttt{pip}-installable Python implementation of our misspecification-aware POPS regression algorithm, following the Scikit-learn \texttt{linear\_model} API~\cite{sklearn}, is provided at \url{https://github.com/tomswinburne/POPS-Regression.git}}. 
    \item \change{We test the approach on synthetic datasets, showing how our scheme provides robust bounds on test errors. This performance is maintained under application to thousand-dimensional datasets for fitting machine learning interatomic potentials\cite{montes2022training,li2020complex} and informatics models\cite{tynes_graphlet_2024,qm9_reference}, untreatable with existing misspecification-aware methods.}
\end{enumerate}

\change{The remainder of this paper is organized as follows. In Section \ref{sec:bayes}, we review Bayesian regression and PAC-Bayes bounds, highlighting the limitations of standard approaches for misspecified models. Section \ref{sec:analysis-ge} presents our analysis of the generalization error for near-deterministic surrogate models, introducing the concept of pointwise optimal parameter sets (POPS). In Section \ref{sec:ansatz}, we develop our misspecification-aware \textit{ansatz}, providing efficient implementations for linear models. Section \ref{sec:numerical_experiments} demonstrates the effectiveness of our approach on both synthetic datasets and high-dimensional problems in atomic machine learning. Section \ref{sec::discussion} discusses our method in the context of existing uncertainty quantification and propagation schemes for atomistic simulation. Finally, Section \ref{sec:conclusion} summarizes our findings and highlights future directions for research in misspecification-aware regression.}

\section{Bayesian fitting of surrogate models}
\label{sec:bayes}
\change{This section summarizes known results from Bayesian analysis and PAC-Bayes inequalities\cite{alquier2021user}, focusing on the limit of interest for this paper: fitting a misspecified surrogate model in the case that the ground truth is (near-)deterministic, i.e. known to have vanishing aleatoric error. We provide a glossary of all terms in Appendix \ref{app:glossary}}
\subsection{Deterministic simulation engines}
A simulation engine $\mE : \mX\to\mY$ takes 
input $\X\in\mX$ and returns $\Y\in\mY$.
\change{We denote the dimension of $\mY$ and $\mX$ by $Y$ and $X$ respectively.}
We consider the regime where the aleatoric uncertainty of $\mE$ is weak,
such that the output $\Y$ follows a Gaussian distribution
\begin{equation}
    \rho_\mE(\Y|\X)
    =
    \mathcal{N}(\Y | \mE(\X),{\bm\Sigma}_\mY(\X)),
    \label{rho_YX}
\end{equation}
which is fully defined by the output mean $\langle\Y|\X\rangle\equiv\mE(\X)$ and covariance
$\langle\Y\Y^\top|\X\rangle
-\mE(\X)\mE^\top(\X)
\equiv
{\bm\Sigma}_\mY(\X)$,
where 
$\|{\bm\Sigma}_\mY(\X)\|=\mathcal{O}(\epsilon^{2Y})$,
and $\epsilon$ is assumed small.
As the aleatoric error is vanishingly weak, we assume homoskedatiscity 
${\bm\Sigma}_\mY(\X)={\bm\Sigma}_\mY$ for simplicity. We will later take
the deterministic, underparametrised limit $\epsilon\to0$, $N/P\to\infty$ assuming that $\epsilon=\mathcal{O}(\sqrt{P/N})$.
With weights $\sum_{i=1}^N{\rm w}_i=1$,
the training data $\mD_N=\{\Y_i,\X_i\}_{i=1}^N$
form an input distribution in $\X$ of
\begin{equation}
    \rho_{N}(\X)
    \equiv\sum_i{\rm w}_i\delta(\X-\X_i)
    ,\,\,
    \rho_{\mD}(\X)\equiv
    \lim_{N\to\infty}\rho_{N}(\X).
    \label{rho_x}
\end{equation}
Learning requires that we can bound expectations over independent identically distributed (i.i.d.) samples $\X,\Y\in\mD$, 
where $\mD$ is the set of all possible input-output pairs, 
corresponding to the limit of infinite training data, 
\begin{equation}
    \langle f(\Y,\X) \rangle_{\mD}
    \equiv
    \int_\mD
    f(\Y,\X)
    \rho_\mE(\Y|\X)\rho_{\mD}(\X)
    \rd\X
    \rd\Y.
    \label{expectation_f}
\end{equation}
For timeseries or other signal data the i.i.d. property is only satisfied for decorrelated segments $\X = \{{\bf x}_{t}\}_{t=1}^T$, meaning 
${\bm\Sigma}_\mY$ should capture signal correlations as in 
generalized least squares regression\cite{alquier2021user,germain2016pac,masegosa2020learning}.\\
\subsection{Deterministic surrogate models}
A surrogate model $\mM : \mX\to\mY$ for $\mE$ 
is defined by $P$ parameters $\T\in\mP$.
Under a parameter distribution $\pi(\T)$, 
the predicted distribution has the same form 
as (\ref{rho_YX}), reading
\begin{align}
    \rho_\mM(\Y|\X,\T)
    &=
    \mathcal{N}(\Y |\mM(\X,\T),{\bm\Sigma}_\mY),
    \\
    \rho_\mM(\Y|\X)&=
    \int_\mP\rho_\mM(\Y|\X,\T)
    \pi(\T)\rd\T.\label{rho_M}
\end{align}
Unlike typical Bayesian schemes\cite{bhat2010derivation,von2011bayesian},
we do not treat the covariance term ${\bm\Sigma}_\mY$
as fitting parameter but as an intrinsic property 
of $\mE$. This has 
implications for model selection criteria,
discussed in appendix \ref{app:model_select}.
\subsection{The generalization error and expected loss}
The generalization error $\mG[\pi]$ is simply the cross entropy of the predicted distribution $\rho_\mM(\Y|\X)\rho_N(\X)$ 
to the observed distribution $\rho(\Y|\X)\rho_N(\X)$; omiting the constant term
$\mG_\mE\equiv\left\langle\ln\left|\rho_\mE(\Y|\X)\right|\right\rangle_\mD$ we have
\begin{equation}
    \mG[\pi]
    =
    -
    \left\langle
    \ln\left|
    \int_\mP\rho_\mM(\Y|\X,\T)\pi(\T)\rd\T
    \right|\right\rangle_\mD.
    \label{G}
\end{equation}
In principle any learning scheme should aim to minimize 
the functional $\mG[\pi]$, but in practice 
learning schemes target a tractable upper bound, the 
\textit{expected loss} $\mL[\pi]\geq\mG[\pi]$,
often known as the negative log likelihood.
Using the Jensen inequality $-\langle\ln x\rangle\leq-\ln\langle x\rangle$, 
\change{as used to derive Gibbs' famous inequality for free energy estimation\cite{gibbs1902elementary}, we find that}
\begin{equation}
    \mathcal{G}[\pi]\leq
    \mL[\pi]
    =-
    \int_\mP
    \left\langle
    \ln\left|
    \rho_\mM(\Y|\X,\T)
    \right|
    \right\rangle_\mD
    \pi(\T)\rd\T.
\end{equation}
$\mL[\pi]$ is clearly minimized by a delta function
$\delta(\T-\T^*_\mL)$, where $\T^*_\mL\in\arg\min_\mP\left\langle
-\ln\left|\rho_\mM(\Y|\X,\T)\right|\right\rangle_\mD$. As the delta function has formally zero width, the loss minimizing distribution thus has vanishing parameter uncertainty. The next section shows how this emerges from the inference posterior at finite $N/P$.
%
%
%
%
\subsection{PAC-Bayes estimation and connection to inference}
\label{sec:pac-bayes}
\change{PAC-Bayes analysis~\cite{alquier2021user} provides 
concentration inequalities~\cite{hoeffding1994probability}
that bound expectations over $\mD$ by empirical expectations over $\mD_N$ 
with some probability.
Whilst originally designed for classification problems,
Germain and coworkers \cite{germain2016pac} established an important connection between Bayesian inference and probability approximately correct (PAC-Bayes) analysis for regression problems with unbounded losses,
which bounds true expectations by empirical likelihood expectations through 
concentration inequalities. A short additional note on the PAC-Bayes framework for regression is provided in appendix \ref{app:pac_constant}.

The PAC-Bayes framework provides a robust learning scheme without 
test-train splitting, and gives a rationalization for Bayesian inference as a 
form of regularization. In particular, minimizing the PAC-Bayes bound for the 
expected loss gives the familiar relation between posterior, prior and 
likelihood from Bayesian inference. This subsection adapts the results of 
Germain \textit{et al.} \cite{germain2016pac} and Masegosa \textit{et al.} \cite{masegosa2020learning} in the near-deterministic limit of interest here. }
The central result of\cite{germain2016pac} is that the expected loss $\mL[\pi]$ admits the PAC-Bayes upper bound, holding with probability $1-\xi$, of
\begin{equation}
\mL[\pi]\leq
\int_\mP
\hat{L}_N(\T)
+
\frac{1}{N}
\ln\left|\frac{\pi}{\pi_0}
\right|
\rd\pi(\T)
-\frac{\ln|\xi|}{N}    
+C_0,
\label{PAC-L} 
\end{equation}
where $\hat{L}_N(\T)=-\langle\ln|\rho_\mM(\Y|\X,\T)\rangle_N
$ is the empirical loss over $\rho_N(\X)$, $\pi_0(\T)$ is some 
\textit{prior} and $C_0$ is a constant.
\change{The value of the constant $C_0$ requires certain assumptions on the distribution of the error between the empirical and true expected loss.
Following \citep{germain2016pac}, we assume that this error distribution is unbounded but sub-normal, i.e. all error fluctuations are bounded 
by a normal distribution of variance $s^2$, giving $C_0[\pi_0]<s^2/2$.
We refer the reader to \citep{germain2016pac} and subsequent works 
\citep{shalaeva2020improved,masegosa2020learning} for further discussion.
The minimizer of the upper bound (\ref{PAC-L}) 
is the posterior from Bayesian inference \cite{germain2016pac},
reading, with a parameter $\lambda$ to enforce normalization}
\begin{equation}
    \pi_{\mL,N}(\T)=\pi_0(\T)\exp(\lambda-N\hat{L}_N(\T)).
\end{equation}
As $N/P\to\infty$ the posterior is strongly 
peaked around a loss minimizer $\T^*_\mL \in \arg\min_{\T\in\mP} L(\T)$. 
Defining ${\bm\Sigma}^{-1}_0$ as the curvature of $-\ln\pi_0(\T)$ around 
$\T^*_\mL$, we use Laplace's method (appendix \ref{app:laplace}) as 
$N/P\to\infty$ to write
\begin{equation}
    \pi_\mL(\T)=\mathcal{N}(\T|\T^*_\mL,
    [{\bm\Sigma}_0+N\mI^*_\mL]^{-1}),
    \label{rhoT}
\end{equation}
where 
$\mI^*_\mL\equiv
\lim_{N/P\to\infty}
{\bm\nabla}_\T{\bm\nabla}^\top_\T
\hat{L}_N(\T^*_\mL)$ is the Fisher information matrix. 
If $\mI^*_\mL$ is full rank the prior has vanishing 
influence and the expected loss is minimized by the sharp distribution 
$\pi_\mL(\T)\to\delta(\T-\T^*_\mL)$, giving vanishing parameter uncertainty. 
Otherwise, the inference prior may still have finite width but, by construction, these will not have influence on model predictions.  
The generalization error of the 
loss minimizing distribution clearly diverges as $1/\epsilon^2$:
\begin{equation}
    \lim_{N/P\to\infty}
    \mG[\pi_\mL]
    =
    \frac{1}{2}{\rm Tr}({\bm\Sigma^*_\mL}{\bm\Sigma}_\mY^{-1})
    =
    \mathcal{O}(1/\epsilon^2).
    \label{G_L}
\end{equation}
${\bm\Sigma}^*_\mL=\langle[\mE(\X)-\mM(\X,\T^*_\mL)][\mE(\X)-\mM(\X,\T^*_\mL)]^\top\rangle$ is the error covariance around $\T^*_\mL$ and we use ${\rm Tr}({\bm\Sigma}_\mY)=\mathcal{O}(\epsilon^2)$.
Under misspecification it is thus clear that minimizing this upper bound is sub-optimal \cite{masegosa2020learning} as the generalization error diverges as $1/\epsilon^2$. In the next section we analyze the generalization error 
as $\epsilon\to0$ to derive a condition for any $\pi(\T)$ that ensures a finite generalization error. 

\subsection{Related work on misspecification-aware Bayesian regression}
In recent years, multiple groups have used the PAC-Bayes framework to derive tighter bounds for the {generalization error} than that provided by the expected loss ~\cite{masegosa2020learning,germain2016pac,lahlou2021deup,psaros2023uncertainty,lotfi2022bayesian,morningstar2022pacm}, to build a misspecified-aware regression scheme.  However, these efforts focus on probabilistic regression settings, often with neural network models far from the underparametrized limit, meaning aleatoric, epistemic and misspecification errors must be considered jointly.\\

Masegosa \cite{masegosa2020learning} derived second order PAC-Bayes bounds for the generalization error minimized through a variational or ensemble \textit{ansatz}. Morningstar \textit{et al.} \cite{morningstar2022pacm} 
developed estimators to assess disagreement between the generalization error 
and expected loss, deriving specialized PAC bounds for theoretical guarantees.
Lahlou \textit{et al.} \cite{lahlou2021deup} considered misspecification of neural networks, jointly training a minimum loss surrogate and an independent predictor of misspecification error. Lofti \textit{et al.}
\cite{lotfi2022bayesian} considered misspecification 
in the context of deep model selection, deriving conditional likelihoods 
better aligned with the generalization error. As various definitions for the distinction between epistemic and 
misspecification uncertainties exist \cite{kato2022view}, we have defined (section \ref{sec:pac-bayes}) misspecification as non-aleatoric uncertainties 
which survive in the underparametrized limit, where the minimizer of PAC-Bayes bounds for the expected loss, 
which coincides with the posterior from Bayesian inference, 
predicts vanishing parameter uncertainties. 
Our epistemic uncertainties thus coincide with standard estimates 
from loss minimization.

\section{Analysis of the generalization error in the near-deterministic limit\label{sec:analysis-ge}}
We consider the generalization error (\ref{G}) 
in the near-deterministic limit, 
$\mG_0[\pi]\equiv\lim_{\epsilon\to0}\mG[\pi]$.
The integral over $\Y$ in (\ref{G}) concentrates at $\Y=\mE(\X)$; 
using $\langle\dots\rangle_\mX\equiv\int\dots\rho_\mD(\X)\rd\X$ yields, using Laplace's method (appendix \ref{app:laplace})
\begin{equation}
    \mG_0[\pi]
    =
    -\left\langle
    \ln\left|\rho_\mM(\mE(\X)|\X)\right|\right\rangle_\mX
    =
    -
    \left\langle
    \ln\left|
    \int_\mP\rho_\mM(\mE(\X)|\X,\T)\pi(\T)\rd\T
    \right|
    \right\rangle_\mX.
\end{equation}
Whilst $\Y=\mE(\X)$ is a unique maximum, to ensure 
$\min_{\X\in\mX}\rho_\mM(\mE(\X)|\X)>0$
and thus avoid divergence in $\mG_0[\pi]$ as $\epsilon\to0$ 
we require 
$\pi(\T)$ has mass in every \textit{pointwise optimal parameter set} (POPS)
\begin{equation}
    \mP_\mM(\X)
    \equiv
    \left\{ \T \, | \mM(\X,\T) = \mE(\X) \right\}.
\end{equation}
Any model with a constant term will ensure $\mP_\mM(\X)\neq\emptyset$, i.e. that the POPS is not empty, but under misspecification it clear that $\cap_{\X\in\mX}\mP_\mM(\X)=\emptyset$, i.e. that no one parameter choice can be a member of all POPS. To express the final generalization error, we introduce a pointwise mass function
\begin{align}
    {\rm m}_\mE(\T,\X)&\equiv\int_\mP
    \delta(\T-\T')
    \rho_\mM(\mE(\X)|\X,\T')\rd\T',
    \label{mass_function}
\end{align}
which gives our first main result, a generalization error for
deterministic surrogate models of
\begin{equation}
    \mG_0[\pi]
    \equiv
    -
    \left\langle\ln
    \int_\mP
    {\rm m}_\mE(\T,\X)
    \pi(\T)
    \rd\T\right\rangle_\mX.
    \label{G0}
\end{equation}
To avoid $1/\epsilon^2$ divergence in $\mG_0[\pi]$ a valid $\pi(\T)$ must satisfy the `POPS covering' constraint as $\epsilon\to0$ of
\begin{equation}
    \min_{\X\in\mX}\max_{\T\in\mP_\mM(\X)}
    \pi(\T)>\lim_{\epsilon\to0}\exp(-c/\epsilon^2), 
    \label{covering}
\end{equation}
for any $c>0$. Equation (\ref{covering}) is our second main result, a demonstration that any candidate minimizer of the generalization error must have 
probability mass in every POPS $\mP_\mM(\X)$, whilst the minimizing distribution concentrates on regions where multiple POPS intersect. 
As shown above, the minimum loss solution $\pi_\mL(\T)$ 
does not satisfy this requirement and thus 
$\mG[\pi_\mL]=\mathcal{O}(1/\epsilon^2)$. 
Importantly, when (\ref{covering}) is satisfied, the predictive distribution $\rho_\mM(\Y,\X)$ covers every training data point, i.e. model predictions will envelope observations as $N/P\to\infty$. 
We now investigate two candidate distributions which satisfy (\ref{covering}),
an $\mathcal{O}(N)$ ensemble \textit{ansatz} and 
$\mathcal{O}(P)$ hypercube \textit{ansatz}, 
finding approximate variational minima that allow for efficient 
deployment on high-dimensional datasets.

%
\section{POPS-constrained ensemble and hypercube \textit{ansatz}\label{sec:ansatz}}
The covering constraint (\ref{covering}) requires $\pi(\T)$
to have mass in the POPS of each training point. This 
condition must be satisfied whilst concentrating mass 
in as small a region in $\mP$ as possible to minimize $\mG_0$.
To achieve this, we first select a set of 
pointwise fits $\T^*_\X\in\mathcal{T}^*_E$
that are each POPS-constrained loss minimizers
\begin{equation}
    \T^*_\X\in\mathcal{T}^*_E,\quad
    \T^*_\X \equiv \arg \min_{\T_\X\in\mP_\mM(\X)}
    L(\T_\X).
    \label{POPS-constrained}
\end{equation}
These pointwise fits will be closely clustered 
around the minimum loss solution $\T^*_\mL$, 
as measured by the Fisher metric $\mI^*_\mL$.
In appendix \ref{app:ensemble} we explore a weighted ensemble \textit{ansatz} 
\begin{equation}
    \pi^*_E(\T)=\langle{\rm w}(\X)\delta(\T-\T^*_\X)\rangle,
\end{equation}
where ${\rm w}(\X)>0$. In particular, we show
$\mathcal{T}^*_E$ naturally emerges when choosing pointwise fits $\T_\X\in\mP_\mM(\X)$to minimize the \textit{ansatz} expected loss
$\mathcal{L}[\pi_E]=\langle{\rm w}(\X)L(\T_X)\rangle$.\\

The ensemble \textit{ansatz} produces 
highly informative max/min bounds for any test point, 
providing an envelope for worst 
case errors, as we show in figure \ref{fig::polynomial}.
However, even when employing the variational optimal reweighting 
${\rm w}^*(\X)$ (see appendix \ref{app:variational-ensemble}), 
the ensemble \textit{ansatz} systematically underestimates moments of 
the test error distribution, as shown in figure \ref{fig::test_errors}.
In addition, both resampling and storage are $\mathcal{O}(N)$, 
which can become problematic for the $N/P\to\infty$ limit of interest. 
As a result, whilst further investigation of ensemble schemes 
and alternatives to the variational optimal reweighting ${\rm w}^*(\X)$ (appendix \ref{app:variational-ensemble}), we now look for an alternative \textit{ansatz} beyond the ensemble approach. 

For practical applications, the parameter distribution should ideally 
require at most $\mathcal{O}(P^2)$ effort for storage and $\mathcal{O}(P)$ for 
resampling. We have found this can be achieved by finding the minimal hypercube 
$\mathcal{H}(\mathcal{T}^*_E)\in\mathcal{P}$
which encompasses all members of the loss minimizing 
POPS-ensemble $\T^*_\X\in\mathcal{T}^*_E$. 
As $\mathcal{H}(\mathcal{T}^*_E)$ intersects all POPS by construction, 
equation (\ref{covering}) is satisfied by the 
hypercube \textit{ansatz} uniform over 
$\mathcal{H}(\mathcal{T}^*_E)$, i.e.
\begin{equation}
    \pi^*_\mathcal{H}(\T)
    =
    V^{-1}_\mathcal{H}\delta(\T\in\mathcal{H}(\mathcal{T}^*_E)),
\end{equation}
where $V_\mathcal{H}=\int_\mP 
\delta(\T\in\mathcal{H}(\mathcal{T}^*_E)){\rm d}\T$.
Crucially, resampling $\pi_\mathcal{H}(\T)$ requires only $\mathcal{O}(P)$ effort
whilst giving a conservative estimate and bounding of test errors. 

To determine $\mathcal{H}(\mathcal{T}^*_E)$ we use singular value decomposition 
on the POPS-ensemble $\mathcal{T}^*_E$ to obtain the right eigenmatrix 
${\bf V}\in\mathbb{R}^{R\times P}$, $R\leq P$, ${\bf V}{\bf V}^\top=\mathbb{I}_R$, 
requiring $\mathcal{O}(P^3)$ effort for evaluation.
Uniformly sampling the full max/min hypercube of the 
projected ensemble $\tilde{\mathcal{T}}^*_E=\{{\bf V}\T|\T\in{\mathcal{T}^*_E}\}$ returns samples $\tilde{\T}^\top{\bf V}$ of $\pi^*_\mathcal{H}$ for $\mathcal{O}(R)$ effort and 
$\mathcal{O}(RP)$ storage. We find this 
efficient hypercube posterior to give excellent bounding and prediction 
of test errors for $P\in[3,1600]$, as shown in figures \ref{fig::test_errors} and \ref{fig::md_ratio}. 

\subsection{POPS-constrained loss minimization for linear models}
Both the ensemble and hypercube ansatz require POPS-constrained loss minimizing fits $\T^*_\X\in\mathcal{P}_\mM(\X)$. Practical application 
therefore requires an efficient means to perform this constrained minimization. Linear models use feature functions 
$\mathcal{F}:\mX\to\mP\times\mY$
to map inputs $\X\in\mX$ to features $\F(\X)\in\mP\times\mY$,
giving $\mM(\X,\T) = \T\cdot\F(\X)$ with parameters $\T\in\mP$. 
Minimizing the PAC-Bayes bound (\ref{PAC-L}) with a Gaussian prior $\pi_0(\T)=\mathcal{N}(\T|{\bf0},{\bm\Sigma}_0)$ gives a global loss minimizer
$\T^*_\mL$ of
\begin{equation}
    {\bf A}=
    [{\bm\Sigma}_\mY{\bm\Sigma}^{-1}_0/N+\langle\F(\X)[\F(\X)]^\top\rangle]^{-1}_\mX
    ,\quad
    \T^*_\mL=
    {\bf A}\langle\F(\X)\mE(\X)\rangle_\mX
    \label{min_loss},
\end{equation}
The epistemic covariance for small $N/P$ can be incorporated in a variational 
PAC-Bayes scheme (appendix \ref{app:variational-ensemble}) that will be developed further in future work. However, in the following, we concentrate on misspecification uncertainties in the limit $N/P\to\infty$, where the influence of $\bm\Sigma_0$ is negligible. For linear models the POPS-constraint is simply $\T_\X\cdot\F(\X)\equiv\mE(\X)$. The loss minimizing pointwise fits $\T^*_{\X}$ can be evaluated analytically via the efficient rank-one updates
\begin{equation}
    \T^*_{\X}
    =
    \T^*_\mL
    +
    \frac{\mE(\X)-\T^*_\mL\cdot\F(\X)}{{\rm h}(\X)}
    {\bf A}\F(\X),
    \label{eq::linearpointwisefits}
\end{equation}
where ${\rm h}(\X)=[\F(\X)]^\top{\bf A}\F(\X)$
is the \textit{leverage}, an outlier measure closely related to the distance of Ref.~\onlinecite{mahalanobis1936generalized}. The minimum loss solution is the leverage-weighted centroid of pointwise optima, $\T^*_\mL=\langle {\rm h}(\X)\T^*_{\X}\rangle_\mX$. 
Equation (\ref{eq::linearpointwisefits}) is our third main result, 
a scheme to incorporate misspecification uncertainty into 
least-squares linear regression for minimal overhead, requiring only $\mathcal{O}(N)$ inner products over feature vectors, which can then be used to construct the ensemble or hypercube \textit{ansatz} given above. \change{Details of the implementation are provided in appendix \ref{app:pseudo-code}. In appendix \ref{app:scaling} we compare the performance of our \texttt{POPSRegression} implementation against \texttt{BayesianRidge} from \texttt{sklearn}, finding an minimal overhead of around $2\times$ for fitting and essentially identical performance for prediction, even when returning max/min bounds in addition to standard deviations.}

\section{Numerical Experiments\label{sec:numerical_experiments}}
\change{In this section we apply our POPS approach to a variety of synthetic and real datasets, comparing our POPS-ensemble and POPS-hypercube \textit{ansatz} to Bayesian inference and the minimization of the PAC-Bayes generalization error $\mathcal{G}_E$. We also discuss our approach in the context of existing methods for uncertainty quantification in atomistic machine learning and discuss how these parameter uncertainties can be propagated up scales to predict material properties with uncertainty\cite{maliyov2024exploring}.}

\subsection{Comparison to Bayesian ridge regression and minimization of $\mathcal{G}_E$}
We applied Bayesian ridge regression and our POPS-\textit{ansatz} to regress a quadratic polynomial against a sinusoidal function, the constant, linear and quadratic features giving $P=3$ model parameters. Appendix \ref{app:envelope}
shows the same procedure for a model with $P=6$ independent features.
For these simple models it is possible to approximately minimize the 
generalization error $\mathcal{G}_E$ of some discrete model ensemble, providing we regularize with a finite aleatoric uncertainty ${\bm\Sigma}_\mY=\sigma{\bm\Sigma}^*_\mathcal{L}$. 
With increasing $\sigma$ we have $\mathcal{G}\to \mathcal{L}$ and thus 
expect the ensemble to concentrate on $\T^*_\mathcal{L}$. 
In the small $\sigma$ regime of interest minimizing $\mathcal{G}_E$ 
becomes increasingly unstable, meaning that the final minimization is highly sensitive to initial conditions and frequently diverges, leading to conditioning issues. We found the same small $\sigma$ instability 
when applying an ensemble approach from Ref.\ \onlinecite{masegosa2020learning} which targets a second-order PAC-Bayes loss bound, designed for misspecified probabilistic models where $\sigma$ can be large. These instabilities show existing misspecification-aware methods require careful hyperparameter tuning and low model dimension to avoid vanishing gradients or other numerical issues. Our POPS-\textit{ansatz}, in contrast, can easily sample $P=\mathcal{O}(1000)$ or higher dimensional data. \\

\begin{figure}[!h]
  \centering
  \includegraphics[width=0.66\columnwidth]{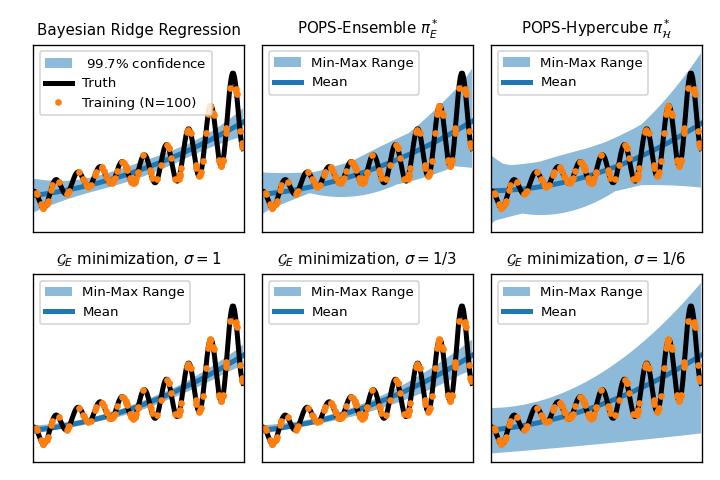}
  \caption{Regression of a deterministic quadratic polynomial ($P=3$) model onto a sinusoidal "simulation engine", trained on $N=100$ points.
  Top left: mean and $3\sigma$ interval from Bayesian ridge regression. All other plots show mean and max/min range. Top: POPS-ensemble and POPS-hypercube \textit{ansatz}
  $\pi^*_E$ and $\pi^*_\mathcal{H}$. Bottom: numerically optimized $\mathcal{G}_E$ for a uniformly weighted $N$-ensemble, regularized with $\Sigma_\mathcal{Y}=\sigma\Sigma^*_\mathcal{L}$, for $\sigma=1,1/3,1/6$. Lower $\sigma$ values gave numerical instabilities and are not presented (see appendix \ref{app:ensemble}).}
  \label{fig::polynomial}
\end{figure}

Figure \ref{fig::polynomial} shows results for $\sigma=1,1/3,1/6$, at the lower limit for which minimization converges. As expected from section \ref{sec:pac-bayes}, Bayesian ridge regression significantly underestimates test errors even at the 99.7\% confidence level (three standard deviations) which as discussed above will worsen with increasing $N/P$. The
max/min range of the numerical ensemble shows similar behavior for $\sigma=1,1/3$ whilst at $\sigma=1/6$ test errors are bounded but significantly overestimated, which worsens 
as $\sigma$ decreases (whilst stable). Importantly, in all cases our POPS \textit{ansatz} $\pi^*_E$ and 
$\pi^*_\mathcal{H}$ give excellent bounding of test errors and 
are stable in the near-deterministic limit $\sigma\to0$. As expected, the hypercube \textit{ansatz} $\pi^*_\mathcal{H}$ gave slightly more conservative bounds. 
In appendix \ref{app:envelope} we provide a similar test of directly minimizing the generalization error for higher dimensional datasets, with $P=6$ independent features. As for the polynomial case, we find that strong regularization is required for minimization to be stable, and in this limit the solution closely follows that of loss minimization. At weaker regularization, the direct minimization becomes systematically closer to the prediction of the POPS ansatz $\pi^*_\mathcal{H}$. These albeit simple results give strong evidence that our \textit{ansatz} is indeed a stable and highly efficient means to find approximate minimizers of $\mG$ that would otherwise be intractable.

\subsection{Test errors in medium-dimensional linear regression}
We now consider higher/medium dimensional problems, $P=10-100$, where 
numerical minimization of $\mG_E$ is not possible. The simulation engines 
were constrained to be misspecified from the linear surrogate models by combining features in a cubic polynomial. We also investigated the influence of adding a simple Gaussian noise term to simulates the presence of variability orthogonal to the chosen feature set.  
In all cases, independent test and training data were generated, 
typically with a 10:90 test:train split. However, all conclusions 
remain robust to varying the precise nature of data generation. In appendix  
\ref{app:envelope} we provide examples showing this performance remains 
when varying the form of simulation engine to quadratic and sinusoidal functions or randomly sampling the linear feature coefficients from some predetermined subset of models.
\begin{figure}[!h]
  \centering
  \includegraphics[width=0.7\columnwidth]{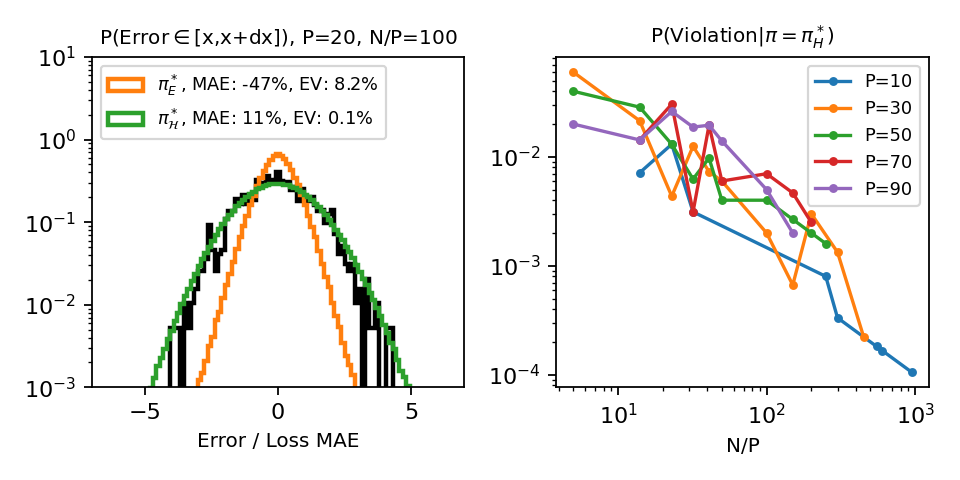}
  \caption{Test errors of a misspecified linear surrogate model on a cubic simulation engine.
  Left: test error histogram at $P=20$, $N/P=100$ for the minimum loss model (black) and predictions from the POPS-ensemble $\pi^*_E$ (orange) and POPS-hypercube $\pi^*_\mathcal{H}$ (green) \textit{ansatz}. MAE: mean absolute error relative to the minimum loss solution. EV: envelope violation, points lying outside of the max/min bound. Right: Probability of envelope violation for the $\pi^*_\mathcal{H}$ \textit{ansatz} with $P$ and $N/P$.}
  \label{fig::test_errors}
\end{figure}

As shown, the $\mathcal{O}(P)$ hypercube \textit{ansatz} 
$\pi^*_\mathcal{H}$ provides an excellent prediction and 
bounding, of the test error distribution whilst 
the $\mathcal{O}(N)$ ensemble $\pi^*_E$ underestimates moments of the error distribution and the max/min envelope bound. 
Importantly, the envelope violation rate of the 
hypercube \textit{ansatz} $\pi^*_\mathcal{H}$ drops with $N/P$, 
as shown in figure \ref{fig::test_errors} for a range of $P$ and $N/P$.
The simultaneous prediction and \textit{bounding} of test errors is 
highly valuable for surrogate model deployment, allowing not only 
the prediction of expected errors but a robust assessment of 
worst case error scenarios. 

\subsection{Application to machine learning interatomic potentials}
We now demonstrate the performance of the method on challenging high-dimensional $P=\mathcal{O}(100-1000)$ regression tasks from atomic machine learning.
Atomic-scale simulations of materials were traditionally broadly classified into two types:
\begin{enumerate}[label=\alph*)]
    \item Empirical simulations with physically-motivated parametric models of interatomic interactions. These allow for fast simulations at a qualitative level of accuracy.
    \item First-principle quantum simulations obtained from approximate solutions of the full Schrodinger equation. These incur a high computational cost and exhibit poor scalability with system size. 
\end{enumerate}
The use of machine learning techniques in atomic simulation has blurred the lines between these two limits by promising near-quantum accuracy at a much more favorable computational cost and scaling \citep{deringer2019machine}. The total energy of the system $V$ is typically factored into a sum of per-atom contributions $V_i$, which are themselves parameterized in term of a feature vector that describes the local atomic environment ${\bf F}_i(\{{\bf x}_j|r_{ij}<r_c\})$, where $r_c$ is a interaction range. 
Training a machine learning model proceeds by minimizing a squared error loss
between reference energies and gradients. While certain approaches are 
systematically improvable, the computational cost associated with repeated 
evaluations can limit the complexity of the models that are selected for 
applications. Further, quantum calculations can be tightly converged, so that 
their intrinsic error (w.r.t. a fully converged calculation) can be made small 
compared to typical errors incurred by computationally-efficient models. This 
regime corresponds to the underparametrized deterministic setting considered 
here. Estimating the errors incurred by such machine learning models is highly 
sought after to assess the robustness of conclusions drawn from the simulations \citep{wen2020uncertainty,li2018uncertainty,gabriel2021uncertainty,wan2021uncertainty}. \change{In the following, we explore the application of 
our POPS approach for a linear interatomic potential model $V_i=\T \cdot
{\bf F}_i$, with forces obtained by taking spatial gradients. Whilst we choose a particlar form of linear interatomic potential model, the presented approach is general to the wide range of available implementations\cite{lysogorskiy2021performant,wood2018extending,podryabinkin2017active,goryaeva2021}.}

\begin{figure}[!h]
  \includegraphics[width=0.7\columnwidth]{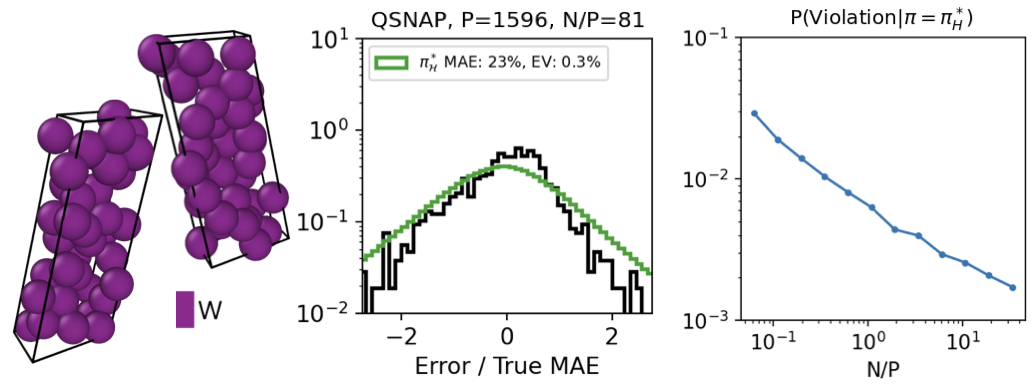}
  \caption{Fitting a \texttt{qSNAP}\cite{wood2018extending} interatomic potential to a diverse tungsten dataset\cite{karabin2020entropy}. Left: representative training configurations\cite{montes2022training}.
  Center: test error histogram at $P=1596$, $N/P=81$ for the minimum loss model (black) and predictions from the POPS-hypercube $\pi^*_\mathcal{H}$ (green) \textit{ansatz}. MAE: mean absolute error relative to the minimum loss solution. EV: envelope violation, points lying outside of the max/min bound. Right: Probability of envelope violation for the $\pi^*_\mathcal{H}$ \textit{ansatz} with $N/P$.
  }
  \label{fig::md_ratio}
\end{figure}

\change{Our first application is to pure tungsten, a key material for nuclear fusion 
applications\cite{goryaeva2021}. Atomic training data was generated using a 
recently introduced information-theoretic approach \citep{karabin2020entropy,montes2022training} 
that aims to maximize an estimated feature entropy for a given set of feature functions, giving an an extremely (ideally, maximally) diverse training set.}
10,000 atomistic configurations containing between 2 and 32 tungsten atoms were 
generated and characterized using Density Functional Theory \citep{kohnsham}, 
yielding 167,922 training data points in total (10,000 energies and 157,922 
gradient components). 
Per-atom energies are expressed as a linear combination of $P=1596$ so-called 
bispectrum components within the `Quadratic' Spectral Neighborhood Analysis 
Potential (\texttt{SNAP}) formalism \citep{wood2018extending}, 
known as \texttt{qSNAP}, which can (in spite of what the name suggests) be expressed in linear model form 
$V_i=\T \cdot{\bf F}_i$ by defining the ${\bf F}_i$ as products of standard \texttt{SNAP} features.
Numerical experiments were carried out by randomly sub-selecting $N$ training points (either energies or gradient components), and by generating a discrete ensemble of $N$ pointwise-optimal models according to Eq.\ \ref{eq::linearpointwisefits}, up to a maximal value of $N=129,853$.\\

The properties of the ensemble are then evaluated on a complementary set of 10,000 hold-out testing points. Error analysis is carried out in the context of the POPS-hypercube \textit{ansatz}. Even on the full dataset, the analysis requires only on the order of 20 minutes on 1 CPU core. It is therefore extremely lightweight in addition to being easily parallelizable, enabling 
its use for extremely large datasets. The results show that the envelope violation on test data varies between $2\%$ and $0.2\%$ for $N/P=0.06$ and $33$, respectively, in qualitative agreement with the results shown in figure \ref{fig::test_errors}. This demonstrates that worst-case model errors are extremely well captured by the POPS-hypercube ensemble. As reported in the left panel of figure \ref{fig::md_ratio}, these excellent bounds do not trivially result from the resampled error distribution being overly pessimistic, as the distribution of errors induced by a uniform parameter resampling within the POPS-hypercube provides an excellent predictor of the actual error distribution with respect to the minimum-loss model. The MAE estimated from the resampled models exceeds the measured value by only 23\%, while providing a very good description of the overall error distribution.\\

\begin{figure}[!h]
  \centering
  \includegraphics[width=0.7\columnwidth]{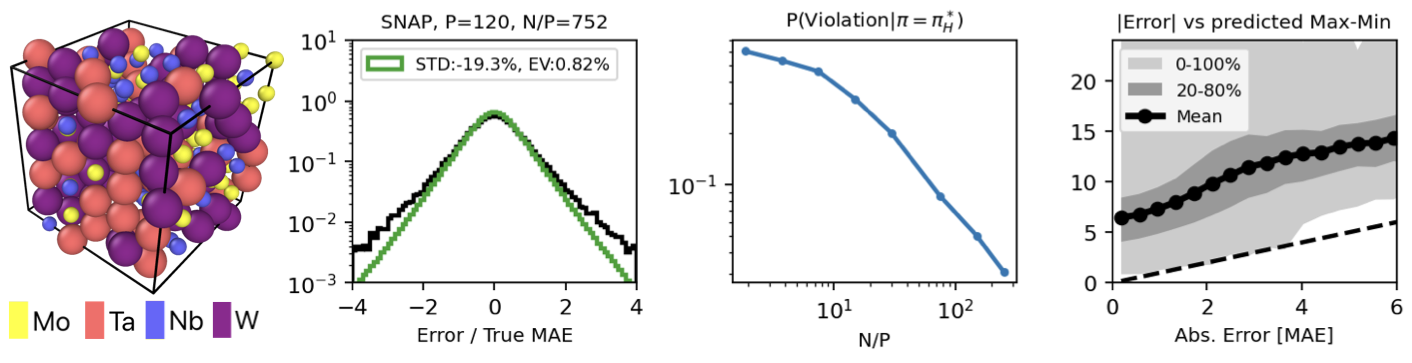}
  \caption{Fitting a \texttt{SNAP}\cite{wood2018extending} interatomic potential to a NbMoTaW high-entropy alloy (HEA) training set\cite{li2020complex}. Far Left: representative HEA configurations. Left:
  test error histogram at $P=120$, $N/P=376$ for the minimum loss model (black) and predictions from the POPS-hypercube $\pi^*_\mathcal{H}$ (green) \textit{ansatz}. MAE: mean absolute error relative to the minimum loss solution. EV: envelope violation, points lying outside of the max/min bound. Right: probability of envelope violation for the $\pi^*_\mathcal{H}$ \textit{ansatz} with $N/P$. Far right: correlation of actual vs predicted error in the test set. See main text.}
  \label{fig::hea_ratio}
\end{figure}

\change{
Our second test case fits a simple \texttt{SNAP} potential to \textit{ab initio} data for the NbMoTaW high-entropy alloy from Li \textit{et al.}\cite{li2020complex}. 
In this case, we use a simple form of multi-component descriptor that is linear in the number of species, and additionally choose a small set of descriptor functions, giving a total of $P=120$ adjustable parameters to fit $N=43,250$ force observations of this chemically complex system, with a further $N=121248$ force observations held out for testing. As can be seen in figure \ref{fig::hea_ratio}, the POPS-hypercube shows excellent performance in bounding prediction errors and in estimating the test error distribution, albeit with a slight underestimation in the tails of the error distribution. 

To investigate the ability of our POPS approach to capture these large-error predictions, we also plot a common metric for uncertainty quantification schemes, the correlation between predicted and observed errors. Whilst such a correspondence is not expected to be exact, as the observed error is a stochastic quantity, there is clear benfit in the ability to asses, with confidence, whether a given prediction is expected to have small or large error. In the present setting, an appropriate comparison is the absolute error of the loss minimizing model against the predicted by the hypercube posterior prediction. A well perfoming model should a) bound the absolute error from above, i.e. not predict low error when the actual error is high and b) have a max-min range which increases for points with large error. As shown in the right of figure \ref{fig::hea_ratio}, we see these properties are indeed observed in the HEA example. An in-depth study of our POPS apporach in active learning schemes will be the subject of future work.}

\subsection{Application to molecular cheminformatics models}
\begin{figure}[!h]
  \centering
  \includegraphics[width=0.7\columnwidth]{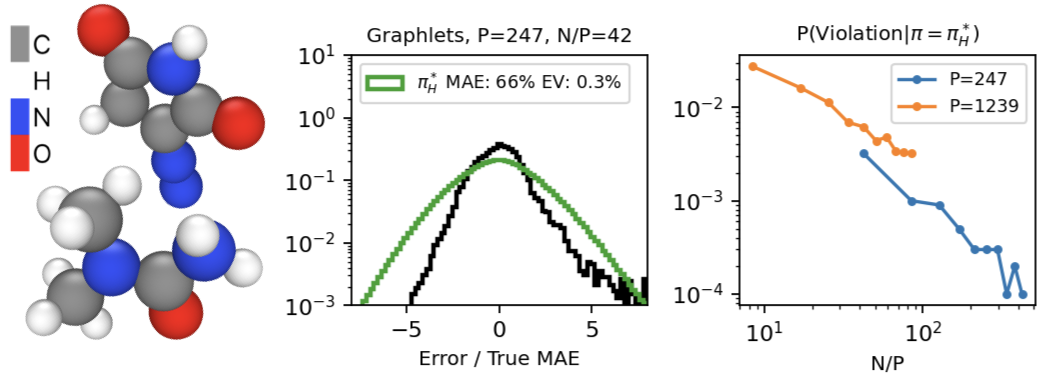}
  \caption{Fitting a linear graphlet model\cite{tynes_graphlet_2024} to predict energies from the QM9 dataset\cite{qm9_reference}. Left: representative small organic molecules. Center: test error histogram at $P=247$, $N/P=42$ for the minimum loss model (black) and predictions from the POPS-hypercube $\pi^*_\mathcal{H}$ (green) \textit{ansatz}. MAE: mean absolute error relative to the minimum loss solution. EV: envelope violation, points lying outside of the max/min bound. Right: Probability of envelope violation for the $\pi^*_\mathcal{H}$ \textit{ansatz} with $N/P$.
  }
  \label{fig::qm9_ratio}
\end{figure}

\change{
As a final application, we apply the POPS scheme to a recently introduced 
linear graphlet model\cite{tynes_graphlet_2024} for predicting molecular properties. Whilst superficially similar to the interatomic potential application, the graphlet features are quite distinct as they are designed to produce 
 explanatory models that detail the contribution of different local bond environment to the total atomic energy rather than providing a smoothly differentiable interatomic potential. 
 The graphlet features, which are essentially discrete counts of the different possible subgraphs of the molecular graph, have a much higher sparsity than the atomic descriptors employed in the previous examples.\\

Here, we consider the formation energies of around 50,000 small organic molecules from the well-known QM9 dataset of small organic molecules\cite{qm9_reference,ramakrishnan2014quantum}. Restricting feature extraction to graphlets of maximum order 3 or 4 gave a total of $P=247$ or $P=1239$ features, respectively. Once again, the formation energy model is linear in the graphlet features.
As shown in figure \ref{fig::qm9_ratio}, the POPS-hypercube again provides an excellent prediction and bounding of test errors, with the probability of envelope violation decreasing with increasing $N/P$. A detailed study of the use of our POPS approach for linear models in a wide range of applications will be the subject of a separate publication.}

\section{Comparison against existing approaches for uncertainty quantification and propagation in atomic simulation\label{sec::discussion}}
\change{As discussed in the previous section, uncertainty quantification schemes for 
machine learning interatomic potentials are actively sought after to assess the 
robustness of conclusions drawn from atomic simulations\cite{wen2020uncertainty,li2018uncertainty,gabriel2021uncertainty,wan2021uncertainty,best_confirmal_2024,kellner2024uncertainty,thaler2023scalable}. Existing approaches for uncertainty 
quantification can be broadly categorized into two types, those that aim to 
directly estimate the error distribution of the surrogate model, and those 
(such as the POPS approach) that aim to capture model errors through the 
determination of parameter uncertainties.

Whilst direct error estimation is central to active learning workflows\cite{Zhenwei-2015,podryabinkin2017active,vandermause2020fly}, these estimates can 
not typically be used to propagate uncertainty through the entire simulation 
engine to predict uncertainty on quantities of interest. This is because the 
result of any atomic simulation workflow is some quantity of interest (e.g. 
formation energies, elastic constants, or relaxed structures). Each quantity of 
interest is an (implicit) function of the parameters of the chosen interatomic 
potential, and will in general vary with the parameters in a non-trivial manner 
due to the strong correlations in the simulation routine. For example, computing defect 
formation energies require comparing two very similar, and thus highly 
correlated simulations. A direct estimate of energy error for each 
configuration will in general significantly overestimate the error in the 
formation energy, as the correlation between the two structures is not 
accounted for.

As a result, schemes which estimate uncertainty in model parameters are
desirable for uncertainty propagation. The propagation can be achieved if 
we can predict how simulations are affected by variation in parameter, either through 
i) implicit differentiation methods\cite{maliyov2024exploring} ii) reweighting of 
existing simulations using e.g. thermodynamic 
perturbation theory\cite{imbalzano2021uncertainty}, or iii) direct resampling 
simulations by drawing samples from the parameter posterior distribution \cite{best_confirmal_2024}. A 
detailed study combining the first two of these approaches (implicit 
differentiation and thermodynamic perturbation) with the POPS procedure will 
be the subject of future work.

The POPS regression procedure presented above is a simple and efficient approach which returns uncertainties on the parameters of misspecified models. 
To the best of our knowledge, existing schemes for uncertainty quantification use as their objective function some form of loss, and thus 
cannot account for misspecification in parameter uncertainties of the model. In this sense, the POPS scheme is unique in its ability to robustly 
capture the parameter uncertainty of misspecified linear models. 

The most directly-comparable scheme to POPS is ensembling, which 
typically train an ensemble of models on random subsets of the training data, 
a process known as bagging\cite{breiman1996bagging}. Various forms of ensemble 
methods are a popular approach to quantify uncertainty in neural network models\cite{abe2022deep} and have been successfully applied to uncertainty quantification for neural network interatomic potentials\cite{imbalzano2021uncertainty,kellner2024uncertainty,vita2024ltau}. 

It is known that bagging and ensemble methods more generally are most effective for high-capacity models in presence of appreciable aleatoric uncertainty\cite{breiman1996bagging}. In the low-noise regime, i.e. with weak aleatoric uncertainty, ensemble uncertainty estimates can be expected to significantly underestimate the true uncertainty, even when using high-capacity models such as neural networks. For low-capacity models, such as the underparametrized linear models considered here, bagging methods are not expected to produce useful uncertainty estimates. In most settings, loss-based uncertainty estimates require significant calibration\cite{imbalzano2021uncertainty,kellner2024uncertainty,vita2024ltau}, or the use of closely-related methods such as conformal prediction\cite{best_confirmal_2024}, which similarly require some form of calibration. In addition, all loss-based schemes should give vanishing parameter uncertainty in the large data limit, as discussed in section \ref{sec:bayes}. Whilst bagging schemes for neural networks do provide 
finite parameter uncertainty, this is predominantly due to the multi-modal loss landscape, and has unclear relationship to the true uncertainty and generalization error of the model. Indeed, even the POPS-ensemble \textit{ansatz} was found to underestimate moments of the error distribution, motivating the hypercube \textit{ansatz}. An interesting direction for future work is to investigate how the POPS coverage criterion, 
equation (\ref{covering}), can be used to systematically 
construct improved ensemble methods in the low-noise regime.

We end this section with a comment on the wider validity of the POPS method for atomic simulation. Clearly, the requirement to be in the underparametrised limit $N\gg P$, can in practice lead to a large amount of training data when $P$ is large. In addition, this condition implicitly assumes that the model is not applied in a strongly extrapolative regime, which is formally true for any ensemble method\cite{kellner2024uncertainty}. Whilst we have not found this to be an issue in practical applications, this theoretical requirement is also common to ensemble methods\cite{kellner2024uncertainty}. 
In practical tests we do not find this to be problematic as the POPS scheme will typically give an overestimate of model uncertainty in regions of low training data density, which is a robust means to quantify extrapolation in high dimension\cite{zeni2022exploring}. Nevertheless, we recommend using one of the many extrapolation grades / outlier measures for high-dimensional linear models\cite{podryabinkin2017active,lysogorskiy2021performant,goryaeva2020dist_score,zeni2022exploring} to ensure that the POPS conditions are satisfied.}

\section{Conclusion\label{sec:conclusion}}
In spite of its ubiquity in practical applications, misspecification is often ignored in Bayesian regression approaches based on expected loss minimization, which can lead to erroneous conclusions regarding parameter uncertainty quantification, surrogate model selection, and error propagation. In this work, we show that the important near-deterministic underparameterized regime is amenable to approximation by two formally simple \textit{ansatz} that exploits the concentration of the cross-entropy 
minimizing ensemble onto pointwise optimal parameter sets (POPS). For linear regression problems, this ensemble can be obtained extremely efficiently using rank-one perturbations of the expected loss solution, from which our optimally weighted POPS-ensemble or POPS-bounding hypercube \textit{ansatz} can be efficiently obtained. Our POPS-constrained 
\textit{ansatz} are shown to produce excellent parameter distributions which give accurate prediction of test error distributions and highly informative worst-case bounds on model predictions, in the important misspecified, near-deterministic regime where existing methods fail. \\

\change{Importantly, POPS encode model error as misspecification uncertainty on model parameters, which is essential for the propagation of uncertainty to simulation results\cite{maliyov2024exploring,imbalzano2021uncertainty}. The application of POPS parameter uncertainties to multi-scale uncertainty propagation will be the subject of forthcoming work. In addition, a clear extension of the POPS approach is to non-linear models, for interatomic potentials\cite{lysogorskiy2021performant,vita2024ltau} or more widely in scientific machine learning. As the loss function is non-convex, the central challenge will be to efficiently generate an optimal set of POPS-constrained loss minimizers, which can be achieved analytically in the linear case. }

\subsection*{Data availability}
\change{A Python implementation of the POPS-constrained
linear regression algorithm, following the Scikit-learn \texttt{linear\_model} API\cite{sklearn}, is available on GitHub at \url{https://github.com/tomswinburne/POPS-Regression.git}, where we also provide a Jupyter notebook to reproduce a selection of results from the paper. The library can also be installed via the \texttt{pip} package manager via \texttt{pip install POPSRegression}.}

\subsection*{Acknowledgements}
{This work was initiated during a visit at the Institute for Pure and Applied Mathematics at the University of California, Los Angeles (supported by National Science Foundation (NSF) grant DMS-1925919) and continued at the Institute for Mathematical and Statistical Innovation, University of Chicago (supported by the NSF grant DMS-1929348). Their hospitality is gratefully acknowledged. 
TDS was supported by ANR grants ANR-19-CE46-0006-1 and ANR-23-CE46-0006-1, IDRIS allocation A0120913455 and an Emergence@INP grant from the CNRS.
D.P. was supported by the Laboratory Directed Research and Development program of Los Alamos National Laboratory under project number 20220063DR.
Los Alamos National Laboratory is operated by Triad National Security, LLC, for the National Nuclear Security Administration of U.S. Department of Energy (Contract No. 89233218CNA000001).}

%

\appendix
\section{Glossary of terms\label{app:glossary}}
Table \ref{tab:glossary} contains a glossary of the key algebraic symbols used in the paper.
\begin{table}[h]
\centering
\begin{tabular}{|c|l|}
\hline
\textbf{Symbol} & \textbf{Description} \\
\hline
$\mE$ & Simulation engine \\
$\mM$ & Surrogate model \\
$\mX$ & Input space \\
$\mY$ & Output space \\
$\X$ & Input vector \\
$\Y$ & Output vector \\
$X$ & Input vector dimension \\
$Y$ & Output vector dimension \\
$\T$ & Model parameters \\
$\mP$ & Parameter space \\
$N$ & Number of training points \\
$P$ & Number of model parameters \\
$\rho_\mE(\Y|\X)$ & Output distribution of simulation engine \\
$\rho_\mM(\Y|\X,\T)$ & Output distribution of surrogate model for given parameter choice\\
$\pi(\T)$ & Posterior parameter distribution \\
$\rho_\mM(\Y|\X)$ & Model output distribution $\rho_\mM(\Y|\X,\T)$ averaged over $\pi(\T)$\\
$\mG[\pi]$ & Generalization error of $\rho_\mM(\Y|\X)$ to $\rho_\mE(\Y|\X)$ under $\pi(\T)$\\
$\mL[\pi]$ & Expected loss of $\rho_\mM(\Y|\X)$ to $\rho_\mE(\Y|\X)$ under $\pi(\T)$\\
$\hat{L}_N(\T)$ & Empirical loss of $\rho_\mM(\Y|\X,\T)$ to $\rho_\mE(\Y|\X)$ with $N$ training points\\
$\mI^*_\mL$ & $X\times X$ Fisher information matrix from empirical loss \\
${\bm\Sigma}_\mY$ & $Y\times Y$ Aleatoric covariance matrix (assumed vanishing, determinant $\mathcal{O}(\epsilon^{2Y})$) \\
$\epsilon$ & Dimensionless measure of aleatoric uncertainty (assumed small)\\
${\bm\Sigma}^*_\mL$ & $Y\times Y$ Error covariance matrix from loss minimization\\
$\pi_0(\T)$ & Prior parameter distribution \\
$\pi_{\mL,N}(\T)$ & Posterior parameter distribution from Bayesian inference \\
$\pi^*_E(\T)$ & Posterior parameter distribution from POPS-constrained ensemble \textit{ansatz} \\
$\pi^*_H(\T)$ & Posterior parameter distribution from POPS-constrained hypercube \textit{ansatz} \\
\hline
\end{tabular}
\caption{Glossary of key algebraic symbols used in the paper}
\label{tab:glossary}
\end{table}

\section{Model selection criteria for deterministic models}
\label{app:model_select}
The Bayesian information criterion (BIC), used to discriminate between surrogate models, is derived by approximating (twice) the negative log evidence as $N/P\to\infty$, assuming a slowly 
varying prior $\pi_0(\T)=\mathcal{O}(1)$. We have
\begin{equation}
    BIC=-2\ln \left|
    \int\exp(-N\hat{L}({\T}))\pi_0(\T){\rm d}\T    
    \right|
    \to
    N{\rm Tr}( {\bm\Sigma}^*_\mL{\bm\Sigma}^{-1}_\mY)
    +
    \ln\left(\|{\bm\Sigma}_\mY\|^NN^P\|\mI^*_\mL\|\right)
    +\mathcal{O}(1).
\end{equation}
We note that the value of the PAC-Bayes loss bound (\ref{PAC-L}) satisfies $BIC=2N\hat{\mathcal{L}}_N[\pi_{\mL,N}]-2C_0$, with $C_0$ model independent, as discussed in Ref.\ \onlinecite{germain2016pac}.
In standard derivations, ${\bm\Sigma}_\mY$ is set to the 
variational maximum likelihood solution ${\bm\Sigma}_\mY={\bm\Sigma}^*_\mL$, 
yielding the familar expression $BIC = P\ln N + N\ln\|{\bm\Sigma}^*_\mL\| + \ln\|\mI^*_\mL\|$, where model independent terms are neglected (principally, terms containing only $N$ and the aleatoric covariance $\Sigma_\mY$).  However, for deterministic regression problems, ${\bm\Sigma}_\mY$ is fixed. Neglecting model-independent terms, this gives an alternative BIC for deterministic models of
\begin{equation}
    BIC = 
    N{\rm Tr}( {\bm\Sigma}^*_\mL{\bm\Sigma}^{-1}_\mY)
    +
    P\ln N
    +\ln\|\mI^*_\mL\|.
\end{equation}
In physical terms, this corresponds to an estimation of a partition function based on an harmonic approximation of the Hamiltonian around its minimum.

\section{Summary of Laplace's method}
\label{app:laplace}
Laplace's method, also known as the steepest descents method, is a well-known identity allowing the evaluation of an integral of an exponentiated function multiplied by a large number. 
The method applies to a function $f(x)$ which is twice differentiable and has a unique maximum in 
some closed interval $[a,b]$ which may be the entire real line. This encompasses all applications in this paper; the extension to multidimensional functions is straightforward. 
The method approximates the integral using a second-order Taylor expansion of $f(x)$ around the maximum of the function, which reduces the problem to a Gaussian integral.
Laplace's method then reads
\begin{equation}
    \lim_{N\to\infty}
    \int_a^b \exp[Nf(x)]\rd{x} = 
    \sqrt{\frac{2\pi}{N}}\sqrt{\frac{\rd^2f(x^*)}{\rd x^2}}^{-1}
    \exp[Nf(x^*)],
    \quad
    x^* = \arg\max_{x\in[a,b]}f(x).
\end{equation}
For further information we refer the reader to e.g. \citep{wong2001asymptotic}.

\section{POPS-constrained ensemble \textit{ansatz}\label{app:ensemble}}
Selecting some set of pointwise fits and 
$\mathcal{T}_E=\{\T_\X\in\mP_\mM(\X)\}_\mX$ 
and weights $\mathcal{W}_E=\{{\rm w}_E(\X)>0\}_\mX$ 
such that $\langle{\rm w}_E(\X)\rangle_\mX=1$, the covering constraint is 
satisfied by
\begin{equation}
    \pi_E(\T|\mathcal{T}_E,\mathcal{W}_E)
    \equiv
    \langle{\rm w}_E(\X) \mathcal{N}(\T|\T_{\X},
    [{\bm\Sigma}_0+N\mI^*_\mL]^{-1})\rangle_\mX,
    \label{ensemble-ansatz}
\end{equation}
For any choice of weights $\mathcal{W}_E$, an
approximate optimal set  $\T^*_\X\in\mathcal{T}^*_E$ can be determined in a variational setting at finite $N$, using (\ref{ensemble-ansatz}) in the PAC-Bayes loss bound 
(\ref{PAC-L}) for the loss $\mL[\pi_E]=\langle{\rm w}_E(\X)L(\T_\X)\rangle_\mX$.
As $P/N\to0$, we show in the sub-section below that
this reduces to the set of POPS-constrained loss minimizers
\begin{equation}
\T^*_\X\in\mathcal{T}^*_E,\quad
\T^*_\X \equiv \arg \min_{\T_\X\in\mP_\mM(\X)}
L(\T_\X)
\end{equation}
The POPS-constrained loss-minimization ensures the fits will be as close as possible 
under the Fisher metric $\mI^*_\mL$, such that $\T^*_\X\to\T^*_\mL$ for specified models. 
Under misspecification, the generalization error of (\ref{ensemble-ansatz}) using $\mathcal{T}^*_E$ has the $P/N\to0$ limit
\begin{equation}
    \mG_0[\pi_E]=
    -\left\langle
    \ln
    \left\langle{\rm w}_E(\X'){\rm m}_\mE(\T^*_{\X'},\X)\right\rangle_{\X'\in \mX}
    \right\rangle_{\X\in\mX},
    \label{ensemble-ansatz-GE}
\end{equation}
where ${\rm m}_\mE(\T,\X)$ is the mass function introduced in (\ref{mass_function}).
In general we expect a uniform weighting ${\rm w}_E(\X)=1$ to be sub-optimal for generalization as the loss-minimizing fits $\T^*_\X$ will be too close to $\T^*_\mL$. 

To optimize ${\rm w}_E(\X)$ we define a density 
$\rho_{\X}^*(\X')={\rm m}_\mE(\T^*_{\X'},\X)/\langle{\rm m}_\mE(\T^*_{\X'},\X)\rangle_{\X'\in \mX}$ of pointwise fits $\T^*_{\X'}$ satisfying
$|\mM(\X,\T^*_{\X'})-\mE(\X)|=\mathcal{O}(\epsilon)$, 
then use Jensen's inequality to write the "loss-like" bound
\begin{align}
\mG_0[\pi_E]
&=\mG_U
-\left\langle
    \ln
    \left\langle{\rm w}_E(\X')\rho_{\X}^*(\X')\right\rangle_{\X'\in \mX}
    \right\rangle_{\X\in\mX}\nonumber\\
&\leq\mG_U
-\left\langle
    \ln
    |{\rm w}_E(\X')|
    \left\langle
    \rho_{\X}^*(\X')\right\rangle_{\X\in \mX}    
    \right\rangle_{\X'\in\mX},
\end{align}
where $\mG_U$ is the generalization error (\ref{ensemble-ansatz-GE}) with 
uniform weighting ${\rm w}_E(\X)=1$. Under the constraint $\langle{\rm w}_E(\X)\rangle_\mX=1$ it is simple to show this upper bound is minimized by
\begin{equation}
    {\rm w}^*_E(\X)\in \mathcal{W}_E^*,\quad 
    {\rm w}^*_E(\X) = 
    \lambda\left\langle
    \rho_{\X'}^*(\X)\right\rangle_{\X'\in \mX}.
    \label{optimal_weights}
\end{equation}
where $\lambda$ ensures 
$\langle{\rm w}^*_E(\X)\rangle_\mX=1$. 
Evaluation requires at most $\mathcal{O}(N^2)$ effort but only
$\mathcal{O}(N)$ for storage and resampling. The final ensemble 
\textit{ansatz} parameter posterior then reads 
\begin{equation}
    \pi^*_E(\T)
    \equiv
    \langle{\rm w}^*_E(\X) \delta(\T-\T^*_{\X})\rangle_\mX.
    \label{optimal_ensemble}
\end{equation}
Equation (\ref{optimal_ensemble}) is a
variationally optimized ensemble \textit{ansatz} for misspecified
regression problems. The worst case of extreme misspecification, 
corresponding to non-intersecting POPS $\mP_\mM(\X')\cap\mP_\mM(\X)=\emptyset$,
gives $\rho_{\X'}^*(\X)=\delta_{\X\X'}$ and thus uniform weights
${\rm w}^*_E(\X)=1$. Interestingly, it is simple to show that optimal weights are also uniform
in the large $\epsilon$  limit, as $\rho_{\X'}^*(\X)$ is then uniform. 
For real systems we find ${\rm w}^*_E(\X)$ is highly non-uniform and
smallest for points near to $\T^*_\mL$, as expected. 
However, while this approaches improves the description of the parameter uncertainty compared to the uniform weight variant described in the main text, 
numerical tests reveal that moments of the predicted ensemble errors are still underestimated, as the "loss like" bounds still bring all points too close to the minimum loss value. This observation (as well as the reduced evaluation cost) motivates the hypercube approach described in the text. However, the max/min bounds of the ensemble, which are insensitive to our choice of ${\rm w}({\bf X})$, remain robust, as shown in figure 1. 

\subsection{Variational optimization\label{app:variational-ensemble}}
For finite $N/P$, we have access to training data $\mD_N=\{\Y_i,\X_i\}_{i=1}^N$ 
to give averages over $\langle\dots\rangle_N$, with some set of 
pointwise fits $\mathcal{T}_E=\{\T_i\}_{i=1}^N$, $\T_i\in\mP_\mM(\X_i)$.
Our ensemble \textit{ansatz} then reads 
\begin{equation}
\pi_E(\T)
=
\langle{\rm w}_{E,i}\mathcal{N}(\T|\T_i,\sigma^2\mathbb{I})\rangle_N
=
\sum_{\X_i\in\mD_N}
{\rm w}_{E,i}
{\rm w}_i
\mathcal{N}(\T|\T_i,\sigma^2\mathbb{I})
,
\end{equation}
where ${\rm w}_{E,i}$ are the discrete ensemble weights and ${\rm w}_i$ 
are the training data weights.

The empirical loss $\hat{\mathcal{L}}_N[\pi_E]$, which contributes 
All $\pi$-dependent terms in the
PAC-Bayes upper bound (\ref{PAC-L}) then read
\begin{equation}
    \hat{\mathcal{L}}_N[\pi_E]
    =
    \int
    \hat{L}_N(\T)\pi_E(\T)
    ,\quad 
    KL(\pi_E|\pi_0)
    =
    \int
    \ln\left|\frac{\pi_E(\T)}{\pi_0(\T)}\right|
    \pi_E(\T)\rd\T.
\end{equation}
To proceed, we define a POPS-constrained variation $\delta_i\T_i$ such that 
$\T_i+\delta_i\T_i\in\mathcal{P}_\mM(\X_i)$, which gives a variation in 
$\pi_E$ of
\begin{equation}
    \delta_i \pi_E(\T)
    \equiv
    \frac{{\rm w}_{E,i}
    {\rm w}_i
    }{\sigma^2}
    \delta\T_i^\top
    \left[
    \T-\T_i
    \right]
    \mathcal{N}(\T|\T_i,\sigma^2\mathbb{I}).
\end{equation}
The upper bound (\ref{PAC-L}) thus has a POPS-constrained variation
\begin{align}
    \delta_i
    \left(
    \hat{\mathcal{L}}_N[\pi_E]
    +
    \frac{1}{N}
    KL(\pi_E|\pi_0)
    \right)
    &=
    \frac{{\rm w}_{E,i}
    {\rm w}_i
    }{\sigma^2}
    \int
    \delta\T_i^\top
    \left[
    \T-\T_i
    \right]
    \left(
    \hat{L}_N(\T)
    +\frac{1}{N}\ln\left|\frac{\pi_E(\T)}{\pi_0(\T)}\right|
    \right)
    \mathcal{N}(\T|\T_i,\sigma^2\mathbb{I})
    \rd\T,\nonumber \\
    &=
    {\rm w}_{E_i}{\rm w}_i
    \delta_i\T_i^\top
    \int
    \nabla_\T
    \left(
    \hat{L}_N(\T)
    +\frac{1}{N}\ln\left|\frac{\pi_E(\T)}{\pi_0(\T)}\right|
    \right)
    \mathcal{N}(\T|\T_i,\sigma^2\mathbb{I})
    \rd\T,
\end{align}
where we use integration by parts. Aside from the POPS constrained variation, this is a standard variational result which in principle allows 
for variational minimization of $\pi_E(\T)$ at finite $N$ for both linear and non-linear models.\\

In the limit $\sigma\to0$ we make the approximation that, for $\T-\T_i=\delta\T$,
$\pi_E(\T_i+\delta\T)\simeq(1/N)\mathcal{N}(\delta\T|{\bf 0},\sigma^2\mathbb{I})$. 
With a Gaussian prior of zero mean and covariance ${\bm\Sigma}_0$ we have 
\begin{equation}
    \int
    \nabla_\T
    \ln\left|\frac{\pi_E(\T)}{\pi_0(\T)}\right|
    \mathcal{N}(\T|\T_i,\sigma^2\mathbb{I})
    \rd\T
    =
    \int
    \left(
    \frac{\T-\T_i}{\sigma^2}
    +
    {\bm\Sigma}_0^{-1}\T
    \right)
    \mathcal{N}(\T|\T_i,\sigma^2\mathbb{I})
    \rd\T
    \to 
    {\bm\Sigma}_0^{-1}\T_i
    ,\quad 
    \sigma\to0.
\end{equation}

In the present work we are primarily 
concerned with the underparametrized, or large data limit $P/N\to0$, where the influence of the prior distribution is negligible, as demonstrated in the main text.  In this regime, 
for small $\sigma$ the variation of the upper bound (\ref{PAC-L}) has a limiting form  of
\begin{equation}
    \delta_i\hat{\mathcal{L}}_N[\pi_E]
    =
    {\rm w}_{E_i}{\rm w}_i\delta_i\T_i^\top\left[ 
    \nabla_\T\hat{L}_N(\T_i)
    +
    \frac{1}{N}{\bm\Sigma}_0^{-1}\T_i\right]
    + \mathcal{O}(\sigma^2).
\end{equation}
This result shows the POPS-constrained variation of the PAC-Bayes loss upper bound converges to 
the gradient of the squared loss evaluated at a POPS-constrained parameter choice, to within a positive multiplicative constant. As a result,
in the underparametrized limit, the variational optimal choice of of POPS-constrained parameter $\T_\X$
will be given by 
\begin{equation}
\T^*_i \equiv \arg \min_{\T\in\mP_\mM(\X_i)}L(\T),    
\end{equation}
in agreement with the result given in the main text.

\section{Implementation of the \textit{ansatz}}
\label{app:pseudo-code}

\begin{algorithm}[h!]
\caption{Find POPS-Constrained loss minimizers}\label{alg:pops-linear-regression}
\begin{algorithmic}[1]
\Require ${\bf X} \in \mathbb{R}^{N \times P}$, ${\bf y} \in \mathbb{R}^N$, ${\bm\Sigma}_0 \in \mathbb{R}^{P \times P}$ (prior covariance)
\State ${\bf C} \gets {\bf X}^\top {\bf X} + \frac{{\bm\Sigma}_0}{N}$ \Comment{Compute regularized feature covariance matrix}
\State ${\bf w} \gets \text{solve}({\bf C}, {\bf X}^\top {\bf y})$ \Comment{Solve ${\bf C}{\bf w} = {\bf X}^\top {\bf y}$ for least squares solution}
\State ${\bf e} \gets {\bf y} - {\bf Xw}$ \Comment{Calculate residuals}
\State ${\bf A} \gets \text{solve}({\bf C}, {\bf X}^\top)$ \Comment{Solve ${\bf C}{\bf A} = {\bf X}^\top$ for influence matrix}
\State ${\bf h} \gets \text{diag}({\bf X} {\bf A})$ \Comment{Compute vector of leverage scores}
\State ${\bf T} \gets {\bf A} \odot ({\bf e} \oslash {\bf h})$ \Comment{Element-wise product and division for POPS-constrained corrections to ${\bf w}$}
\State \Return ${\bf w},{\bf T}$ \Comment{Return the loss minimizer and POPS-constrained corrections}
\end{algorithmic}
\end{algorithm}

\begin{algorithm}[h!]
\caption{Generate POPS-constrained hypercube samples}\label{alg:pops-hypercube}
\begin{algorithmic}[1]
\Require ${\bf T} \in \mathbb{R}^{N \times P}$ (POPS-constrained corrections from Algorithm \ref{alg:pops-linear-regression})
\State ${\bf U}, {\bf S}, {\bf V}^\top \gets \text{SVD}({\bf T})$ \Comment{Perform SVD on T}
\State $R \gets \text{rank}({\bf S})$ \Comment{Determine rank of S}
\State ${\bf V}_R \gets {\bf V}[:, :R]$ \Comment{Keep only the first R right singular vectors}
\State $\tilde{\bf T} \gets {\bf T} {\bf V}_R$ \Comment{Project T onto the singular vectors}
\State ${\bf l} \gets \min(\tilde{\bf T}, \text{axis}=0)$ \Comment{Find min bounds of projections}
\State ${\bf u} \gets \max(\tilde{\bf T}, \text{axis}=0)$ \Comment{Find max bounds of projections}
\For{$i = 1$ to $M$} \Comment{Generate M samples}
    \State ${\bf x} \gets \text{UniformRandom}({\bf l}, {\bf u})$ \Comment{Sample uniformly within bounds}
    \State ${\bf T}_{\text{sample},i} \gets {\bf x} {\bf V}_R^\top$ \Comment{Project back to original basis}
\EndFor
\State \Return ${\bf T}_{\text{sample}}$ \Comment{Return M hypercube samples}
\end{algorithmic}
\end{algorithm}

A slightly modified implementation of this POPS-constrained linear regression algorithm, wrapping the Scikit-learn \texttt{BayesianRidge} class\cite{sklearn}, is available on GitHub at \url{https://github.com/tomswinburne/POPS-Regression.git}. It can be installed (requires Scikit-learn) via pip: \texttt{pip install POPSRegression}.

\begin{figure}[!h]
    \centering
    \includegraphics[width=0.6\columnwidth]{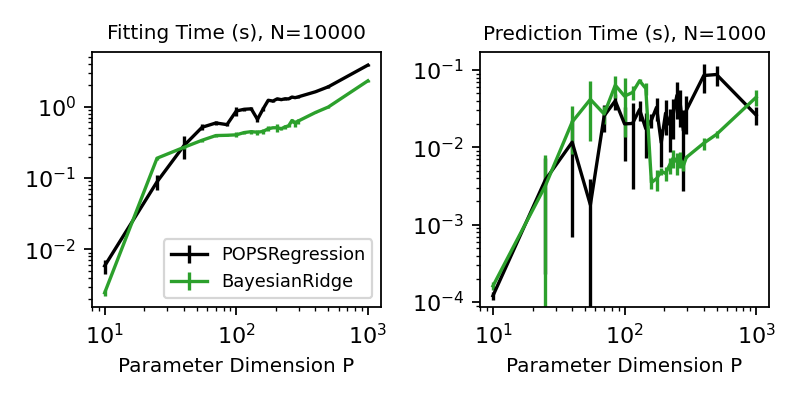}
    \caption{Comparision of \texttt{POPSRegression} against \texttt{BayesianRidge} from \texttt{sklearn}. Left: Mean and standard deviation of the time taken for each method to fit a $P$ dimensional linear model against $N=10000$ training points, with $P$ ranging from 10 to 1000. 
    Right: Mean and standard deviation of the time taken to make $1000$ predictions with mean and epistemic uncertainties for \texttt{BayesianRidge}, whilst \texttt{POPSRegression} returns mean, misspecification uncertainty and max/min bounds. 
    \label{fig::scaling}}
\end{figure}

\section{Performance comparison against Bayesian Ridge\label{app:scaling}}
To demonstrate the performance of \texttt{POPSRegression} against \texttt{BayesianRidge} from \texttt{sklearn}, we fit against the medium-dimensional cubic "simulation engine" described in the main text, with $P$ ranging from 10 to 1000 and $N=10000$. Figure \ref{fig::scaling} shows \texttt{POPSRegression} incurs a minimal ($2\times$) overhead over \texttt{BayesianRidge}, whilst existing misspecification aware schemes do not converge in the low-noise regime. 

\section{Additional plots of test error prediction}
\label{app:envelope}
\subsection{Low dimensional models}
\begin{figure}[!h]
    \centering
    \includegraphics[width=0.6\columnwidth]{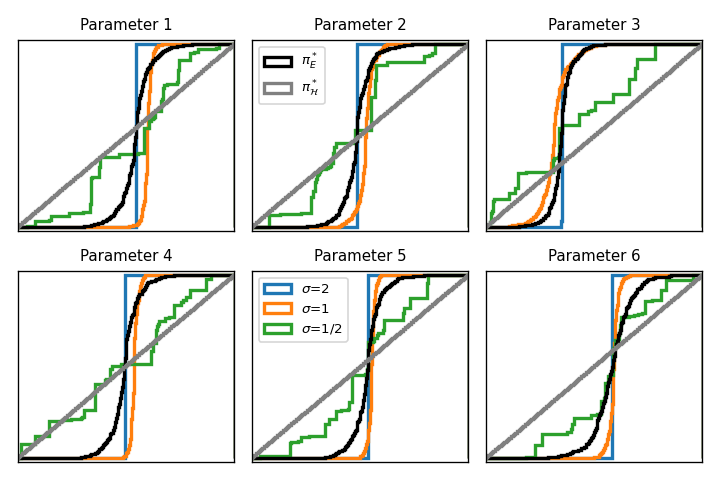}
    \caption{Regression of a linear model onto a cubic "simulation engine" with $P=6$ independent features, trained on $N=500$ points. The panels show the cumulative distribution of each parameter across the $N$-member ensemble. Black, gray : 
    ensemble and hypercube \textit{ansatz} $\pi^*_E$ and $\pi^*_\mathcal{H}$. Blue, orange and green : numerically optimized $\mathcal{G}_E$ for a uniformly weighted $N$-ensemble, regularized with $\Sigma_\mathcal{Y}=\sigma\Sigma^*_\mathcal{L}$, for $\sigma=2,1,1/2$. As in Figure 1 in the main text, lower $\sigma$ gave numerical instabilities.}
    \label{fig::distn}
\end{figure}

We applied the same procedure as for the polynomical example
to a model system with $P=6$ independent features
to simulate more realistic datasets.
As before, the numerical minimization routine could only treat a small range of 
$\sigma$ values. Figure \ref{fig::distn} shows the resulting parameter 
distributions. At large $\sigma=2$, the minimizer quickly concentrates
all ensemble members on the minimum loss solution; at $\sigma=1$ the ensemble 
distribution is notably sharper than the POPS-ensemble $\pi^*_E$, whilst for 
low sigma $\sigma=1/2$ the minimization only has marginal stability, but 
becomes closer to the POPS-bounding hypercube $\pi^*_\mathcal{H}$. In addition, a minimizer initialized with the POPS-ensemble did not change appreciably. 
These albeit simple results give strong evidence that our \textit{ansatz}
is indeed a stable and highly efficient means to find approximate 
minimizers of $\mG$ that would otherwise be intractable.

\subsection{Medium dimensional models}
The plots below are identical in presentation to the top row of figures in the main text, but for the quadratic and random linear simulation engines described in the main text. 
\begin{figure}[!h]
    \centering
    \includegraphics[width=0.49\columnwidth]{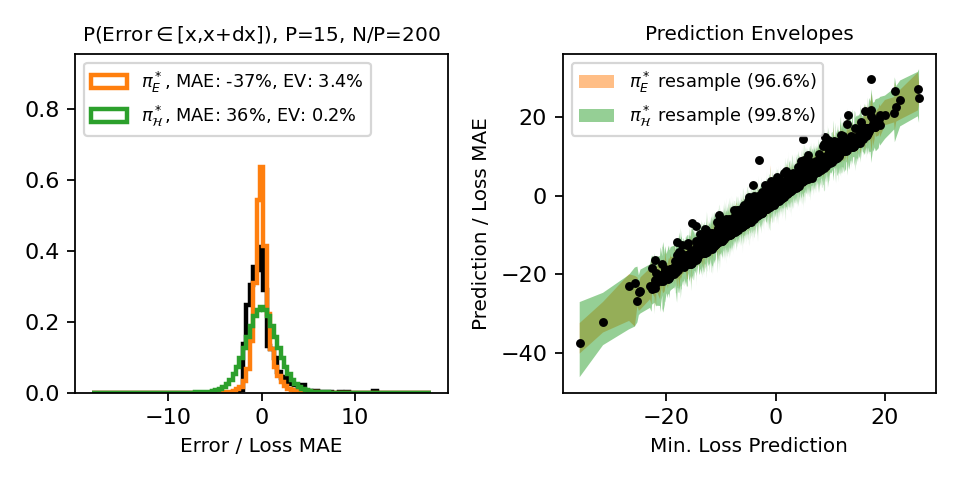}
    \includegraphics[width=0.49\columnwidth]{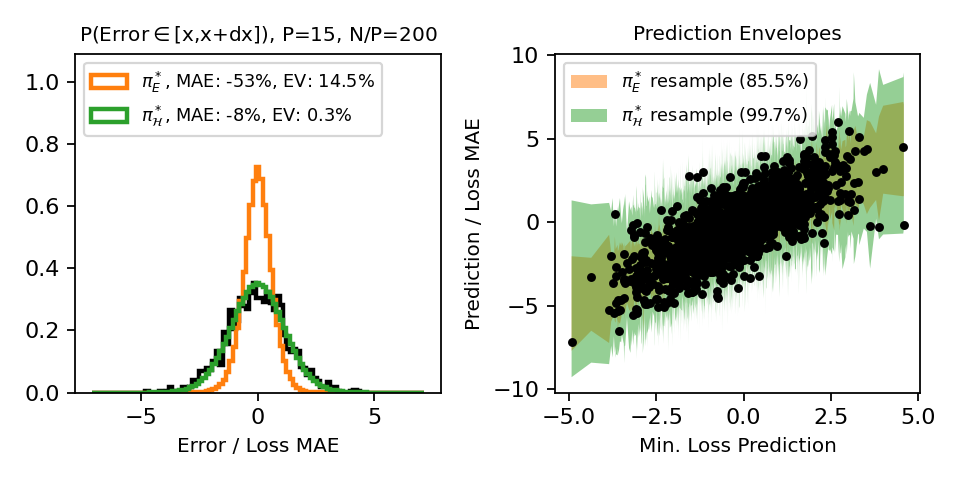}
    \caption{Test error prediction for medium dimensional problems. 
    Left: Observed test errors for a misspecified 
    linear surrogate model (P=15) on a quadratic (top) and random linear (bottom) simulation engine, along with predicted test errors from the POPS-ensemble. 
    Right: Parity plots for the minimum loss model, along with envelope bounds from the POPS-ensemble and hypercube. }
  \end{figure}

\end{document}